# Uncertainty Quantification in Neural-Network Based Pain Intensity Estimation

**Burcu Ozek, Zhenyuan Lu, Srinivasan Radhakrishnan, Sagar Kamarthi**

Mechanical and Industrial Engineering Department, Northeastern University,

Boston, Massachusetts, United States of America

**Abstract**

Improper pain management can lead to severe physical or mental consequences, including suffering, a negative impact on quality of life, and an increased risk of opioid dependency. Assessing the presence and severity of pain is imperative to prevent such outcomes and determine the appropriate intervention. However, the evaluation of pain intensity is a challenging task because different individuals experience pain differently. To overcome this, many researchers have employed machine learning models to evaluate pain intensity objectively using physiological signals. However, these efforts have primarily focused on point estimation of pain, disregarding the inherent uncertainty and variability present in the data and model. Consequently, the point estimates provide only partial information for clinical decision-making.

This study presents a neural network-based method for objective pain interval estimation. It provides an interval estimate for a desired probability of confidence, incorporating uncertainty quantification. This work explores three distinct algorithms: the bootstrap method, lower and upper bound estimation ($Loss_L$) optimized by genetic algorithm, and modified lower and upper bound estimation ($Loss_S$) optimized by gradient descent algorithm. Our empirical results reveal that $Loss_S$ outperforms the other two by providing a narrower prediction interval. It exhibits average interval widths that are 22.4%, 7.9%, 16.7%, and 9.1% narrower than $Loss_L$, and 19.3%, 21.1%, 23.6%, and 26.9% narrower than the results of bootstrap for 50%, 75%, 85%, and 95% prediction interval coverage probability. As $Loss_S$ outperforms, we assessed its performance in three different scenarios for pain assessment: (1) a generalized approach, in which a single model is fit for the entire population, (2) a personalized approach, in which a separate model is fit for each individual, and (3) a hybrid approach, in which a separate model is for each cluster of individuals. Our findings demonstrate that the hybrid approach provides the best performance. Furthermore, the practicality of this approach in clinical contexts is noteworthy. It has the potential to serve as a valuable tool for clinicians, enabling them to objectively assess the intensity of patients' pain while taking uncertainty into account. This capability is crucial in facilitating effective pain management and reducing the risks associated with improper treatment.

# 1. Introduction

Failing to treat pain properly can have serious consequences. Untreated pain can significantly impact the quality of life and cause physical or mental suffering. Inappropriately managed pain can result in over-prescription or under-prescription. While over-prescription may lead to opioid dependency and drug-seeking behavior, under-prescription can cause suffering, that is otherwise avoidable [1,2]. The key to preventing untreated or inappropriately managed pain is to assess its presence and severity to decide on the required intervention [3]. Clinicians employ various pain assessment methods, including the Visual Analog Scale (VAS), where patients indicate their pain level on a 10cm line (0 indicates the absence of pain, while 10 denotes the most severe pain imaginable), the Verbal Rating Scale (VRS) using descriptive terms such as 'none' to 'excruciating, or the Numeric Rating Scale (NRS) from 0 to 10 [4,5]. They must understand the pain scale since accurate and efficient pain measurement aids in early diagnosis, disease monitoring, and evaluating therapeutic efficacy [6].

Even though the VAS, NRS, or VRS methods are easy to deploy in clinical settings, accurately assessing pain intensity can be challenging because pain is subjective and varies based on individual physiological, emotional, cognitive, and communication differences [7]. For instance, one person may find a certain level of pain mildly irritating, another may find the same pain excruciating. Moreover, there are limitations to verbally expressing the pain level in certain patient groups (such as non-verbal children) and medical situations (such as sedative medication or during paralysis) [8]. These limitations may result in poor pain management and potential harm. Automated and objective pain intensity assessment has gained popularity among machine learning researchers to address these issues. Machine learning models can learn from data by automatically detecting and using patterns to predict pain intensity or derive new insights [9].

In literature, many researchers studied machine learning models such as logistic regression, decision trees, support vector machines (SVR), and neural networks to assess pain intensity objectively. These algorithms are developed to learn from facial expressions, body movements, electrodermal activity, electrocardiogram, and electromyogram collected from individuals [1,10-16]. This data, however, is subject to noise and uncertainty due to factors such as motion or muscle artifacts, natural environmental conditions (temperature fluctuations, background noise), changes in skin resistance due to sweating or hydration levels, and individual differences [17,18]. Additionally, machine learning models encounter challenges such as inherent randomness, hyperparameter settings, model assumptions, and complexity [19,20]. These challenges cause an inevitable uncertainty [21,22].

At present, objective pain intensity assessment research focuses only on point estimation, disregarding the variability in the data, uncertainty in the model, or both [23-26]. Point estimations tend to produce overconfident predictions. Overconfident incorrect predictions can be harmful in clinical settings [27]. Understanding the level of uncertainty in pain intensity predictions is critical. It can be achieved by capturing and expressing the inherent uncertainty in the model inputs and parameters, which is then conveyed through the model to quantify uncertainty in the model outcomes [24].

Neural network (NN)-based prediction intervals (PIs) methods are generally recognized as effective for quantifying uncertainty [23,24,26,28]. A PI is an interval estimate for an (unknown) target value [25]. In contrast to a point estimate, a PI includes the lower and the upper bound within which the actual target value is expected to lie with a pre-defined probability [25,28,29].

The two metrics assess the quality of the PIs: (1) *accuracy*, which is represented by Prediction Interval Coverage Probability (PICP), and (2) *dimension*, which is quantified by Prediction Interval Width (PIW). In literature, four traditional methods, namely delta, Bayesian, bootstrap, and mean-variance estimation-based, are employed to create NN-based PIs [23,30,31]. These methods, however, demand high computational resources or make strong assumptions about the data or the model [28]. Additionally, their primary goal is to maximize PICP, but not to minimize PIW. A 100% PICP can be obtained by setting a large value for the upper bound and a small value for the lower bound of PIs. However, this approach provides no useful information about the target value. However, in practice, to achieve high-quality PIs, one must maximize PICP and, at the same time, minimize PIW[25].

In this work, we explore pain intensity interval estimation by implementing three distinct neural network-based models: the bootstrap method, lower and upper bound estimation ($Loss_L$) optimized by genetic algorithm, and modified lower and upper bound estimation ($Loss_S$) optimized by gradient descent algorithm. Our findings reveal that the $Loss_S$ approach consistently outperforms the other two models by providing narrower intervals. We then conduct a comprehensive analysis of the applications of the $Loss_S$ approach across three distinct scenarios: (1) a generalized approach where only one model is fitted for the overall population, (2) a personalized approach where separate models are tailored for each individual, and (3) a hybrid approach where individual models are created for clusters of individuals. To our knowledge, this study is the initial effort to develop a prediction interval method for uncertainty quantification in the field of pain intensity estimation.

The rest of the paper is organized as follows. The background section explores the objective pain assessment concept and prediction interval framework. The methods section introduces the data and the process of constructing PIs. The results and discussion section presents various NN-based PI estimation approaches and assesses their applications in different scenarios. The conclusion section highlights the results and limitations of this work and provides insights into the prospective trajectory of the work.

## 2. Related Work

This section consists of two subsections. The first subsection reviews recent studies in automated pain assessment that have concentrated on leveraging biomarkers and machine learning models for point estimation. The second subsection discusses the introduction of prediction interval techniques to accurately capture and represent the inherent uncertainty in the models and data.

### 2.1 Pain Assessment

Automated and objective pain assessment has gained increasing interest among machine learning researchers in the literature over the years [1,32]. Researchers report that physiological signals, such as brain activity, cardiovascular activity, and electrodermal activity, are a rich source of information for developing objective pain assessment methods. These signals are connected to the autonomic nervous system and play an essential role in pain response[21,33]. Because of these reasons, pain researchers have widely used physiological signals to develop objective and automated pain assessment methods [1,5,8,10-16].

One of the most well-known datasets of pain-related physiological signals, is "The BioVid Heat Pain Database," created by Walter et al. [34]. This dataset consists of electrodermal activity (EDA), electrocardiogram (ECG), electromyogram (EMG), and electroencephalography (EEG). EDA, which measures the skin's electrical properties (skin conductance), is a valuable indicator of neurocognitive stress [17,35]. ECG analyzes the electrical activity related with the heart [36]. EMG is the measurement of muscle activity [33]. EEG measures the brain's electrical activity [37]. Researchers report that EDA is one of the most valuable signals for automated and objective pain assessment [1,16,17,38-43]. In the literature, several researchers used EDA signals from BioVid Dataset to develop different machine learning models with low root mean squared error (RMSE).

Kächele et al. [43] employed a random forest algorithm on EDA signals from the BioVid Dataset; they reported the best performance RMSE as 1.01. Martinez et al. [16] applied linear regression, supper vector regression, neural network, fully-connected recurrent neural networks, and long short-term memory networks to the EDA signals. They achieved RMSE of 1.36, 1.37, 1.32, 1.29, and 1.29, respectively. Pouromran et al. [1] explored linear regression (RMSE: 1.18), SVR (RMSE: 1.15), neural networks (RMSE: 1.15), random forest (RMSE: 1.15), KNN (RMSE: 1.17), and XGBoost (RMSE: 1.13) on BioVid EDA signals.

All of these studies developed point estimation algorithms that do not take into account the uncertainty in the data or model [23-26]. It is crucial to include uncertainty to avoid potential harm, particularly in clinical settings. Neglecting uncertainty may result in either over or under treatment. While over-treatment and over-prescription of opioids can increase the risk of addiction and overdose, the opposite can negatively impact mental and physical health, reduce the quality of life, necessitate longer hospital stays, and cause dissatisfaction with the treatment [27,44-46].

## 2.2 Prediction Interval Framework

Prediction, which plays a crucial role in decision-making, is highly susceptible to any source of uncertainty affected by input data, measurement errors, model approximation errors, parameter uncertainty, and model bias [24,47,48]. Causes of uncertainty in the prediction framework are grouped into (1) model uncertainty (epistemic uncertainty) and (2) irreducible variance (data noise variance or aleatoric uncertainty) [28]. The uncertainty framework in prediction is formulated as follows:

$$\sigma_y^2 = \sigma_{model}^2 + \sigma_{noise}^2 \tag{1}$$

When predicting, the impact of uncertainty term $\sigma_y^2$ should not be overlooked. Researchers have commonly used NN-based PIs to identify and analyze uncertainty. A PI is a forecast of the upper and lower bounds between which the unknown future value of the target $y = f(x)$ is expected to lie, with a specified confidence level, $(1 - \alpha)\%$ [23,24,30,31,49]. In literature, researchers have used two criteria to evaluate the quality of a PI: (1) Prediction Interval Coverage Probability (PICP), and (2) Prediction Interval Width (PIW). PICP is the probability that estimated PIs will cover the actual target value [23,30,50,51]. While a high PICP can easily be achieved by setting a wide PI, it has no predictive value [23]. Hence, consideration of Mean Predication Interval Width (MPIW), the average distance between the estimated lower and upper bounds, is necessary for creating high-quality PIs [23,30,50,51].

Having narrow PIs (i.e., small MPIW) with high PICP is the most desirable outcome in practice. Thus, it is valid to say that there is a tradeoff between these two criteria when evaluating the quality of PIs. A higher desirable PICP could lead to a wider MPIW, and vice versa. [23,52]. Researchers have proposed several methods in the literature for constructing NN-based PI estimation models that address the tradeoff between PICP and MPIW. In literature, there are four traditional methods to build NN-based PI estimation models: (1) bootstrap, (2) delta, (3) mean-variance estimation (MVE), and (4) Bayesian [28,30]. However, they all face common disadvantages: they demand high computational resources and make strong assumptions about the model or input data. The bootstrap method is one of the most frequently employed techniques. It involves constructing a specific number (B) of NN models by resampling the training data from the original data with replacement. The outputs of NN models are averaged to estimate the actual regression mean. The output of NNs also calculates the variance of predictions. The resulting mean and variances are used for constructing the PI. This method has the following drawbacks: (1) it is computationally expensive when dealing with large datasets, and (2) it could provide inaccurate estimations due to bias when the observation set is small or not representative [28,30,31,53].

Khosravi et al. [23] developed a new approach referred to as the Lower Upper Bound Estimation (LUBE), to overcome the limitations of the aforementioned traditional NN-based PI estimation methods. In LUBE, the NN model has two output neurons, one for the upper bound and the other for the lower bound of the PIs. The parameters of the NN model (biases and weights) are optimized considering a novel loss function. This loss function does not directly minimize the regression error; instead, it aims to improve MPIW and PICP simultaneously. Khosravi et al. [23] employed a simulated annealing (SA) method to optimize this novel loss function, considering it nonlinear, complex, discontinuous, and non-differentiable. LUBE is more reliable than traditional techniques and does not require any assumptions about the data or model distributions.

Since the LUBE method performs well and does not impose restrictions on the data distribution and model structure, many researchers adapted it by utilizing various evolutionary optimization algorithms. Quan et al. [54] used the LUBE for electrical load forecasting and optimized it using particle swarm (PSO). Lian et al. [31] adapted the LUBE method to generate NN-based PIs for the landslide displacement; they combined particle swarm optimization and gravitational search algorithm (GSA) to optimize the neural network. Shen et al. [55] developed a multi-objective artificial bee colony algorithm (MOABC) incorporating multi-objective evolutionary knowledge (EKMOABC), and optimized wavelet neural network to create PIs with the LUBE for wind power forecasting.

Although LUBE performs well, researchers reported some limitations. When PIs are zero, the loss function finds its global minimum at 0 [28]. The loss function is highly nonlinear, complex, discontinuous, and non-differentiable and hence only evolutionary algorithms (e.g., simulated annealing and particle swarm optimization could optimize this loss function. These algorithms require a slow training process, and the gradient descent algorithm (GD), the standard neural network training technique, struggles to optimize this particular loss function effectively [31].

To overcome these challenges, researchers made improvements to the LUBE function and modified the loss function or treated it as a multi-objective optimization problem **[56-62]**. Khosravi et al. [58] proposed the concept of independent width and penalty factors as a solution; they employed additive terms instead of multiplication. Quan et al. [54] modified the loss function by improving the interval width assessment metric; they explored the LUBE loss function's PIW based on the mean absolute percentage error as well as the mean squared error principle, which penalizes more when models produce larger error terms. This change aims to overcome the global minimum problem when the width is zero. Secondly, they used an additional loss function to consistently assess the quality of PIs using the scoring rule called SCORE developed by Winkler et al. [63], which rewards a narrow PI and penalizes the cases where the target value is not within the PI. Lian et al. [31] proposed a single hidden layer feedforward NN, which required optimization of only the output layer weights (hidden layers' weights are chosen randomly) to make the training process more efficient. In addition, they trained the model with a hybrid evolutionary algorithm, integrating PSO with GSA. Lastly, they modified the LUBE loss function's PIW part by defining combinational coverage width-based criterion with one-norm regularization. Ak et al. [25] and Shen et al. [55] approached this problem as a multi-objective optimization problem. Ak et al. [25] utilized a multi-objective genetic algorithm, the non-dominated sorting genetic algorithm–II (NSGA-II), to construct PIs. Shen et al. [55] introduced the EKMOABC technique to train the network.

To tackle the convergence towards a global minimum when PIW is zero and to ensure the differentiability and compatibility with GD, Pearce et al. [28] introduced modifications. They replaced the step function of PICP (1 when the actual target value is inside the PI, otherwise 0) with a differentiable approximation by incorporating a softening factor and sigmoid function. Next, they replaced the conventional PIW definition with a captured PIW approach. This change involves calculating PIW only for data points where the actual value falls within the upper and lower bounds of the PI. By considering solely the captured data, the potential influence of non-captured data on the calculation of PIW was mitigated. Furthermore, they modified the impact of PIW in the loss function by transforming multiplicative terms to additive terms. Similarly, the effect of PICP was altered by replacing the exponential term with the squared term. Lastly, they added additional hyperparameters to provide a high confidence in PICP (see Section 3 for details).

In this study, we present an NN-based PI method by integrating two distinct loss functions, called $Loss_L$ and $Loss_S$, which are discussed in detail in the next section. Subsequently, we compare the PIs generated by these loss functions with those constructed using the bootstrap method, which serves as the baseline for the evaluation. The first loss function, $Loss_L$, explicitly targets the challenges associated with traditional NN-based methods, which often require significant computational resources and rely on strong assumptions about the model or input data for uncertainty quantification in pain intensity estimation. To address the limitation of $Loss_L$ function, namely the convergence towards incorrect optimal solutions in specific scenarios, we adopted the second loss function, $Loss_S$. By exploring $Loss_L$ and $Loss_S$ along with the comparative bootstrap method, we aim to address these challenges and improve the accuracy and robustness of pain intensity estimation within the uncertainty quantification framework.

# 3. Methods

## 3.1 BioVid Heat Pain Database and Feature Extraction

In this work, we trained the models using the BioVid Heat Pain Database, a publicly available dataset, to construct PIs for pain intensity estimation [34]. This database includes (1) physiological modalities (EDA, EEG, ECG, and EMG) and (2) behavioral modalities (facial expression) of 87 participants. This data set's target variable (label) is the pain intensity, ranging from 0 to 4. In the BioVid experiments, each participant was exposed to four different temperature levels (T1, T2, T3, and T4), which were personalized for each subject. The absence of temperature-induced stimuli was considered the control temperature level (T0). In the current work, we consider pain intensity level as a continuous variable, and the modeling is considered as a regression task.

In this study, we utilized the EDA signal, which is widely accepted as a neurocognitive stress indicator in pain recognition research [1,17,35,64]. We extracted features from the EDA using the "Canonical Time-series Characteristics" defined by Lubba et al. [65]. These features consist of basic statistical measures of time-series data, stationarity, entropy, linear correlations, physical nonlinear time-series analysis techniques, linear and nonlinear model parameters, predictive power, and fits [65]. In this work, we used the 22 most informative features identified by Lubba et al.[65] (see Appendix S1). We used all these 22 features because our exploration demonstrated that models constructed with these features consistently outperformed those created with different feature combinations.

After the preliminary data cleaning procedures, such as missing value analysis, the dataset contained 8612 observations, each with 22 features. These features were standardized via min-max normalization. The label assigned to each instance corresponds to the level of pain intensity, which varies between 0 to 4.

This study assessed three distinct scenarios and generated corresponding prediction intervals (PIs). First, we created a "generalized" model that is subject-independent and applicable to all subjects covered in the study. Second, we developed 87 "personalized" models, one for each subject. Third, we grouped the 87 subjects into 4 clusters based on their EDA features. For each subject, we constructed 110-dimensional vectors (22 features x 5 pain intensity levels) using the mean of the normalized features in each pain level. We then applied the k-means clustering technique to group the individuals based on subjects' EDA signals. Cluster-specific models were built for each cluster, which is a "hybrid" of the generalized and personalized approaches [1]. All the models were trained using 90% of the observations and evaluated with the remaining 10% of the observations with a ten-fold cross-validation scheme.

## 3.2 Development of Prediction Intervals by Neural Network

Section 3.2.1 delves into the neural network structure, Section 3.2.2 introduces the evaluation metrics used to evaluate the prediction intervals' quality, and Section 3.3.3 and Section 3.3.4 provide detailed information about the loss functions employed to optimize the prediction intervals.

### 3.2.1 Neural Network Structure

Our study employed an NN-based PI model to assess the uncertainty of predictions. The network consisted of one input layer with 22 neurons (each representing a distinct EDA feature) and two hidden layers. The hidden layers varied from 10 to 120 neurons, depending on the scenario. The output layer utilized two neurons: one for the lower bound and the other for the upper bound of PI."

The hidden layer utilized the rectified linear unit (ReLU) function as an activation function, while the output layer employed the linear function as an activation function. The neural network architecture, including the number of hidden layers, the number of neurons in each hidden layer, and the choice of activation functions, were optimized through the hyperparameter tuning process. **Table 1** illustrates the search space for these parameters.

*Table 1: The hyperparameter tuning process optimizes neural network parameters using their search spaces.*

| Parameter | Search space |
|---|---|
| # of Hidden Layers | [1,4] |
| # of Hidden Neurons | [10,150] |
| Activation Function for Hidden Layers' Neurons | [ReLU, Hyperbolic Tangent, Linear] |

### 3.2.2 PI assessment

We evaluated the constructed PIs' quality by employing PICP and MPIW measures. We aim to create PIs as narrow as possible (i.e., PIs with small MPIW) with a PICP as high as possible. PICP is calculated by:

$$PICP = \frac{1}{n}\sum_{i=1}^{n} k_i \quad (2)$$

where *n* is the number of observations,

$$k_i = \begin{cases} 1, & \text{if } L(X_i) \leq X_i \leq U(X_i) \\ 0, & \text{else} \end{cases} \quad (3)$$

where, $L(X_i)$ is the lower bound, $U(X_i)$ is the upper bound of the PI of the $i^{th}$ observation.

Then, we calculated MPIW as:

$$MPIW = \frac{1}{n}\sum_{i=1}^{n} U(X_i) - L(X_i) \quad (4)$$

We calculated the normalized mean prediction interval width (NMPIW) as follows:

$$NMPIW = \frac{MPIW}{R} \tag{5}$$

where $R$ represents the range of the target, $R = \max(y) - \min(y)$, and NMPIW represents the prediction interval relative to the target range.

### 3.2.3 $Loss_L$

We employed the loss function from LUBE to evaluate the BioVid Heat Pain Database [23]. In this work, we refer to LUBE's loss function as $Loss_L$, which is calculated by the following:

$$Loss_L = \frac{MPIW}{R}(1 + \gamma(PCIP)e^{-\eta(PICP-\mu)}) \tag{6}$$

where $R$ represents the range of the target variable, which, in this application, is the pain intensity measured on a 0-4 scale; $\mu$ and $\eta$ are constant hyperparameters; $\mu$ which represents the confidence level associated with PIs, which can be set to $1 - \alpha$; $\eta$ amplifies any small discrepancy between PICP and $\mu$. The term $\gamma(PCIP)$ is a step function that evaluates the PIs on the test set. For training, $\gamma(PCIP)$ is considered as 1 [23], where

$$\gamma = \begin{cases} 0, & PICP \geq \mu \\ 1, & PCIP < \mu \end{cases} \tag{7}$$

We trained the NN model with a Genetic Algorithm (GA), a search heuristic inspired by the natural evolution theory [66]. It repeatedly changes initial solutions by choosing individuals from the current population as parents and uses them to build the children for the next generation at each stage. As the population evolves with each iteration, the algorithm can find the optimal solution [67,68]. We implemented GA using the PyGad Python Library developed by Gad [69]. **Figure 1** shows the process of GA.

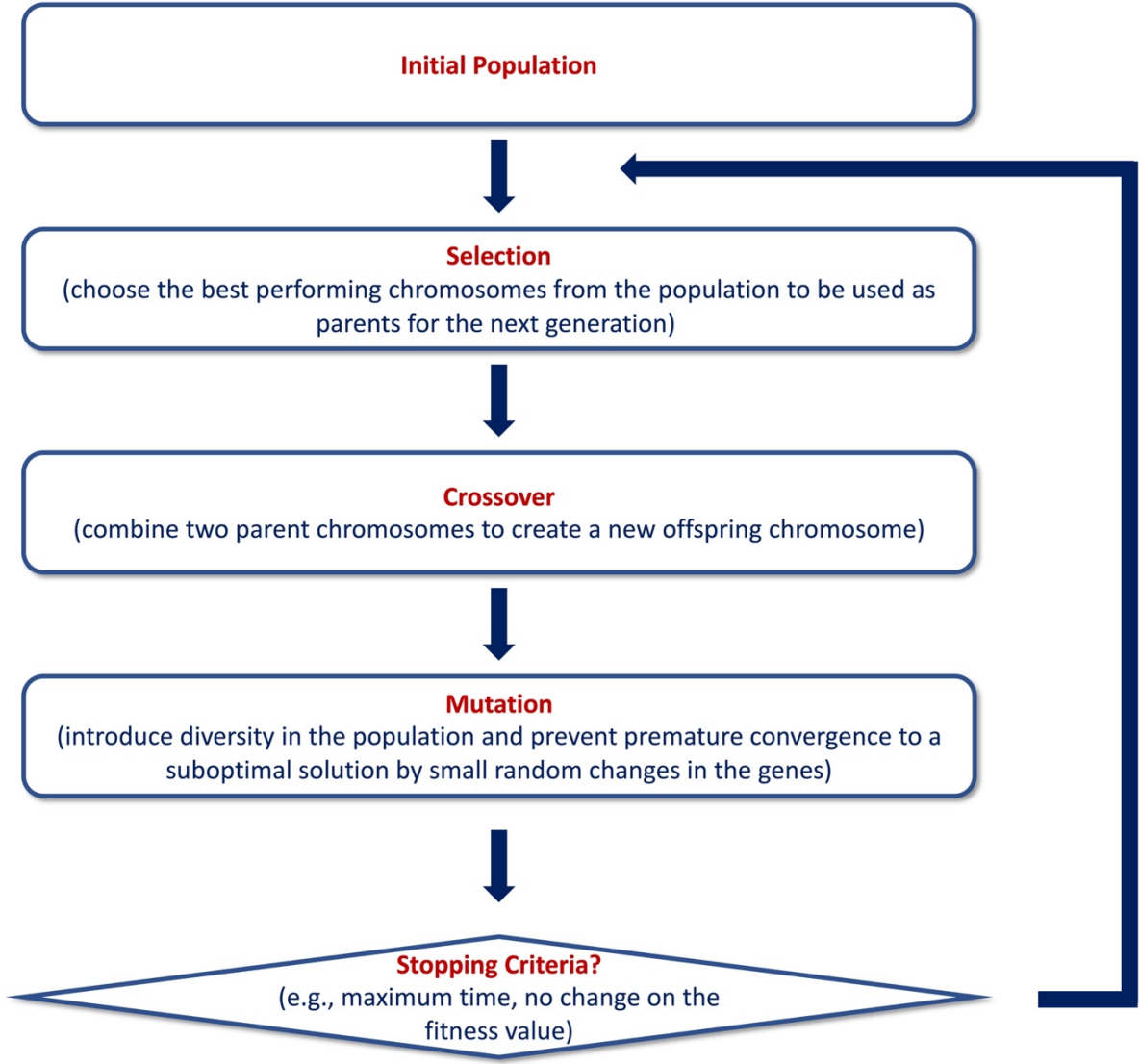

*Figure 1: GA consists of selection, crossover, and mutation steps.*

**Table 2** shows the parameters of GA and Loss$_L$ and their search space for hyperparameter tuning. The optimal values of the parameters were found by hyperparameter tuning.

*Table 2: GA and loss function (Loss$_L$) parameters are optimized using their search spaces through the hyperparameter tuning.*

| Parameter | Search space |
|---|---|
| # of Solutions (GA parameter) | [10,20] |
| # of Parent Mating (GA parameter) | [5,10] |
| % Genes (GA parameter) | [10,20] |
| $\eta$ (loss function parameter) | [25,100] |
| $\mu$ (loss function parameter) | [0.5,0.95] |

### 3.2.4 $Loss_S$

We implemented a modified version of LUBE's loss function and explored its performance on BioVid data [28]. In the modified version, the formulas for MPIW and PICP were changed. We refer to the modified LUBE's loss function as soft loss function, denoted as $Loss_S$. We use $MPIW_S$ and $PICP_S$ to refer to MPIW and PICP, respectively.

The soft loss function $Loss_S$ is calculated by:

$$Loss_S = MPIW_S + \lambda \frac{\eta}{\alpha(1-\alpha)} \max(0, (1-\alpha) - (PICP_S)^2) \tag{8}$$

where

$$PICP_s = \frac{1}{n} \sum_{i=1}^{n} \sigma(s(X_i - L(X_i))) \otimes \sigma(s(U(X_i) - X_i)) \tag{9}$$

$$MPIW_S = \frac{1}{\sum_{j=1}^{n} k_j} \sum_{i=1}^{n} k_i [U(X_i) - L(X_i)] \tag{10}$$

In the above expressions, $\sigma$ is the sigmoid function; $\otimes$ represents matrix multiplication; $s$ is the softening factor $\lambda$ is Lagrangian to control the impact of $MPIW_S$ and $PICP_S$; $\eta$ is a constant hyperparameter that represents the batch size; $\mu = 1 - \alpha$ represents the confidence level associated with PIs; $MPIW_S$ captures MPIW only when $U(X_i) \leq y_i \leq L(X_i)$ holds; $PICP_S$ calculated by replacing the step function, $k_i$, with a smooth a sigmoid function.

$Loss_S$ is differentiable and compatible with the gradient descent (GD) training; GD is an iterative optimization algorithm that identifies the local minimum of a function. The algorithm calculates the gradient of the objective function and adjusts the model's parameters in the opposite direction of the gradient [70-73]. The detailed steps of GD are as follows:

1. Start by randomly initializing the parameters for the model, i.e., NN parameters.
2. Compute the loss function ($Loss_S$).
3. Compute the gradient of the loss function regarding the parameters, which corresponds to the first-order derivative of the function at the local point (slope at the local point).
4. Take a step in the opposite direction of the gradient, and move towards the minimum of the loss function. This step is taken by multiplying the gradient by a scalar value called the learning rate ($\xi$) and subtracting the results from the current parameter values.

$$\theta = \theta - \xi \frac{\partial J(\theta)}{\partial \theta} \qquad (11)$$

where θ is the current NN parameter values (weights); ξ is the learning rate; J(θ) is the loss function calculated by the θ; $\frac{\partial J(\theta)}{\partial \theta}$ is the gradient of the loss function in regard to the current parameter values.

5. Update parameters with the new values obtained in the previous step.
6. Repeat steps 2-5 until the gradient becomes close to zero or a stopping criterion is met; a gradient close to zero indicates that the parameters have converged to a minimum of the loss function.

The optimal values of the parameters and Loss$_S$ and GD algorithm were found by hyperparameter tuning. **Table 3** shows the search space of the parameters.

*Table 3: GD and soft loss function (Loss$_S$) parameters are optimized using their search spaces through the hyperparameter tuning.*

| Parameter | Search space |
| --- | --- |
| Learning rate (GD parameter) | [0.001 ,0.1] |
| Decaying rate (GD parameter) | [ 0.000001, 0.0001] |
| $\lambda$ (loss function parameter) | [5,30] |
| $\eta$ (loss function parameter) | [35,240] |
| $\mu = 1 - a$ (loss function parameter) | [0.5,0.95] |
| $s$ (loss function parameter) | [10,220] |

# 4. Results and Discussion

Section 4.1 compares the performance of the PIs of the generalized models built using the bootstrap method (baseline model), Loss$_S$ optimized by GD, and Loss$_L$ optimized by GA. Section 4.2 discusses the performance of PIs constructed using Loss$_S$ optimized by GD for the generalized, personalized, and hybrid models.

## 4.1 A Comparative Analysis: NN-Based PIs versus Bootstrap, Loss$_L$ by GA, and Loss$_S$ by GD

This section provides a comparative analysis of PIs generated by Loss$_S$, Loss$_L$, and bootstrap. The NN-based models were trained using the EDA signals of all 87 subjects. The goal is to construct PIs with the maximum probability of coverage and minimum width. We used 22 features extracted from the EDA signals and considered pain intensity level as the response variable, a continuous value ranging from 0 to 4.

**Figure 2** illustrates the PICP and MPIW values for bootstrap, Loss$_L$ by GA, and Loss$_S$ by GD methods. The findings demonstrate that Loss$_S$ outperforms the others. Specifically, Loss$_S$ exhibits MPIWs that are 22.4%, 7.9%, 16.7%, and 9.1% narrower than the results of Loss$_L$, and 19.3%, 21.1%, 23.6%, and 26.9% narrower than the results of bootstrap, for PICP values of 50%, 75%, 85%, and 95% respectively.

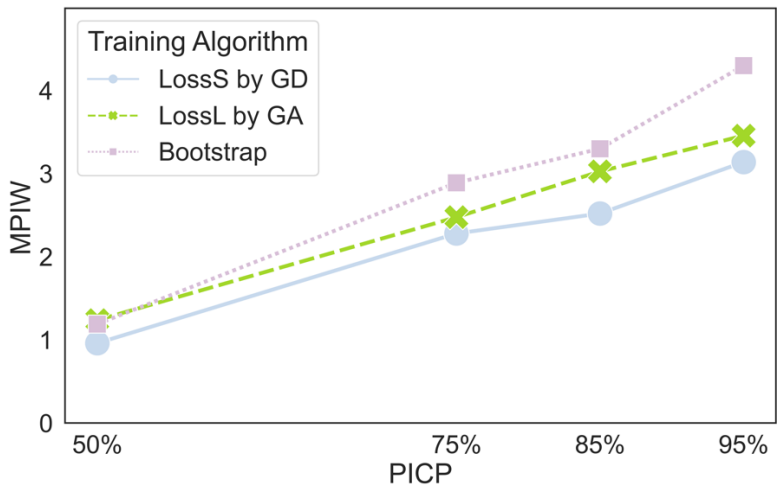

*Figure 2: Performance comparison of Loss$_S$ by GD, Loss$_L$ by GA, and bootstrap methods. Loss$_S$ outperforms by yielding a narrower PIW across all PICP values.*

For example, when considering a coverage probability of 75%, Loss$_S$ yields a PIW of 2.3, whereas Loss$_L$ produces a PIW of 2.5, and bootstrap constructs a PIW of 2.9. Similarly, at a coverage probability of 85%, Loss$_S$ generates a PIW of 2.5, Loss$_L$ constructs a PIW of 3.0, and bootstrap constructs a PIW of 3.3.

The results presented in **Figure 2** indicate that there exists a direct connection between the PICP and MPIW. For example, in the Loss$_S$ method, with a coverage probability of 85%, the MPIW is approximately 2.52. This result means that on average, when the pain intensity level is 3 on a scale of 0 to 4, the estimated range, with an 85% coverage probability, typically spans from 1.31 to 3.51. When aiming for a higher coverage probability (95%), the MPIW increases accordingly. For instance, a pain intensity level of 3 on a scale of 0 to 4 results in the PI of 0.59 and 3.89 on average.

To attain a more comprehensive understanding of the relation between MPIW and PICP, comprehending the impact of Loss$_S$ function hyperparameters on the training process is essential. In **Figure 3**, we observe how the hyperparameters of Loss$_S$, $\lambda$ (the Lagrangian constant that determines the relative importance of MPIW$_S$ and PICP$_S$) and $s$ (the softening factor, which relaxes the original PICP definition) affect the MPIW$_S$ and PICP$_S$. **Figure 3a** focuses on the effect of $s$ while keeping all other parameters constant, and **Figure 3b** examines the impact of $\lambda$ while keeping all other parameters constant. When we compare different $s$ values at the same MPIW$_S$ level, we see that higher $s$ values generally result in better PICP$_S$ values. This means an increase in $s$ value yields higher coverage probability with a narrower PIW. For example, for an MPIW$_S$ range between 1.5 and 2, $s$ values smaller than 110 result in a PICP$_S$ between 50% and 60%, but with a $s$ values larger than 110, we can achieve PICP$_S$ higher than 0.75, most of the time. **Figure 3b** shows an increasing trend, indicating that higher λ values result in slightly higher PICPs.

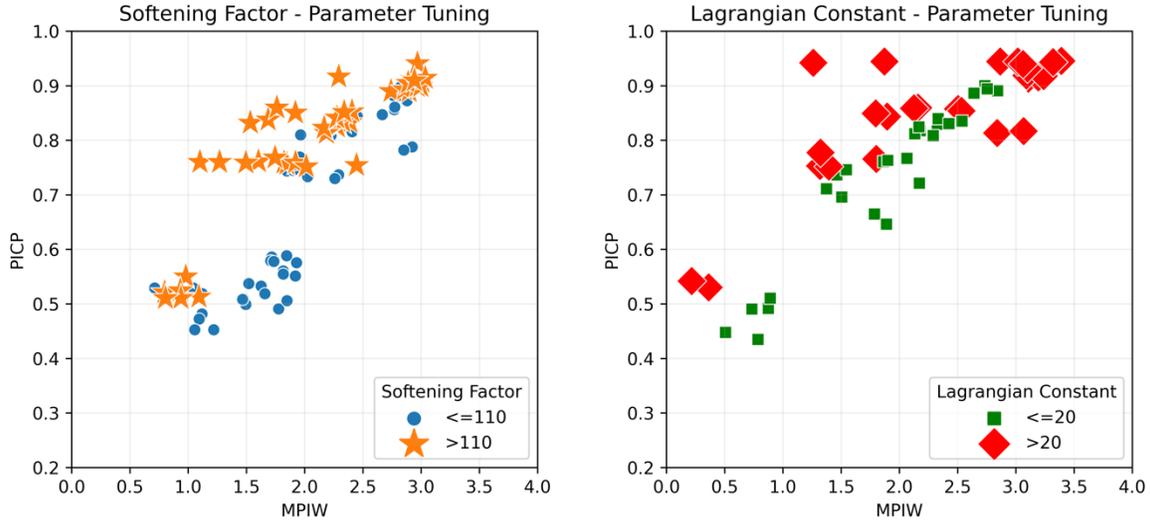

*Figure 3: (a) higher softening factor (s) value improves the PICP. (b) A higher Lagrangian multiplier (λ) value results in a slightly higher PICP.*

### 4.2 NN-Based PI Estimation with Loss$_S$ by GD Algorithm

As the NN-based PI model trained with the Loss$_S$ and optimized by GD algorithm outperforms the Loss$_L$ optimized by GA and bootstrap for the generalized model, we investigated its application for two other scenarios: personalized model and hybrid model. This section provides more detailed information about each of these scenarios.

#### 4.2.1 Generalized Model

We used the EDA signals of all 87 subjects to develop generalized models that are applicable to the entire population. **Table 4** presents the results of the generalized models for various coverage probabilities: 50%, 75%, 85%, and 95%. **Table 5** presents the mean of the upper and lower bounds for each pain level as PICP varies.

*Table 4: The generalized model results demonstrate how MPIW and NMPIW change as PICP varies.*

| PICP | MPIW | NMPIW |
|------|------|-------|
| 50%  | 0.96 | 0.24  |
| 75%  | 2.28 | 0.57  |
| 85%  | 2.52 | 0.63  |
| 95%  | 3.14 | 0.79  |

*Table 5: The mean of the upper and lower bounds for each pain level in the generalized model as PICP varies.*

| | GENERALIZED MODEL | | | | | | | |
|---|---|---|---|---|---|---|---|---|
| | 50% PICP | | 75% PICP | | 85% PICP | | 95% PICP | |
| Target | Lower Bound | Upper Bound | Lower Bound | Upper Bound | Lower Bound | Upper Bound | Lower Bound | Upper Bound |
| 0 | 1.32 | 2.09 | 0.17 | 2.47 | 0.21 | 3.12 | 0.06 | 3.28 |
| 1 | 1.48 | 2.32 | 0.31 | 2.9 | 0.51 | 3.37 | 0.16 | 3.45 |
| 2 | 1.68 | 2.75 | 0.56 | 2.99 | 0.74 | 3.47 | 0.31 | 3.5 |
| 3 | 2.15 | 3.27 | 1.13 | 3.35 | 1.31 | 3.51 | 0.59 | 3.68 |
| 4 | 2.83 | 3.77 | 2.23 | 3.82 | 2.1 | 3.84 | 1.12 | 3.89 |

As an illustration, when considering a 75% coverage probability, the MPIW measures approximately 2.28. This outcome signifies that, on average, for a pain intensity level of 3 on a 4-point scale, the estimated range typically extends from 1.13 to 3.35. While the constructed PIs exhibit reasonable performance, the key strength lies in the generalizability of the proposed approach. With this method, as new patients arrive, their pain intensity prediction intervals can be constructed without requiring additional model training, which has significant implications for clinicians who seek to objectively evaluate the pain intensity levels of their patients instead of relying solely on self-reported information for pain treatment and management.

### 4.2.2 Personalized Model

We developed personalized models, which were custom-trained for each subject using only the subject's own data. In this approach, 87 different personalized models were created for every 87 subjects. **Table 6** shows the averaged PICP, MPIW, and NMPIW values across 87 personalized models. **Table 7** presents the mean upper and lower bounds for each pain level, derived from averages of the personalized models.

*Table 6: Compared to generalized models' results, the PI widths of the individualized model are wider.*

| PICP | MPIW | NMPIW |
|---|---|---|
| 50% | 1.63 | 0.41 |
| 75% | 2.44 | 0.61 |
| 85% | 2.89 | 0.72 |
| 95% | 3.12 | 0.78 |

*Table 7: The mean upper and lower bounds for each pain level averaged across participants for various PICP values.*

|  | PERSONALIZED MODEL | | | | | | | |
|---|---|---|---|---|---|---|---|---|
|  | 50% PICP | | 75% PICP | | 85% PICP | | 95% PICP | |
| Target | Lower Bound | Upper Bound | Lower Bound | Upper Bound | Lower Bound | Upper Bound | Lower Bound | Upper Bound |
| 0 | 0.05 | 2.04 | 0.28 | 2.75 | 0.04 | 2.96 | 0.01 | 3.2 |
| 1 | 0.6 | 2.4 | 0.38 | 2.93 | 0.21 | 3.17 | 0.07 | 3.42 |
| 2 | 0.5 | 2.51 | 0.63 | 3.15 | 0.34 | 3.36 | 0.41 | 3.66 |
| 3 | 1.58 | 2.7 | 0.99 | 3.48 | 0.71 | 3.61 | 0.8 | 4 |
| 4 | 2.01 | 3.03 | 1.49 | 3.72 | 1.42 | 4.04 | 1.2 | 4.19 |

In comparison with the generalized model's findings, the PIWs are larger for personalized models. This is mainly because personalized models have a very limited number of observations for training, making it difficult for the models to learn. In addition to the poor performance, the lack of generalizability makes it unsuitable for clinical settings. In the case of a new patient arriving at the hospital with no patient history, a new model would need to be developed and trained with the patient-specific EDA observations, which may not be feasible in a clinical setting. Nonetheless, this approach can help build smart, personalized devices that can collect vast amounts of data from individuals and use this data to train and customize models for each individual.

### 4.2.3 Hybrid Model

In a clinical setting, personalized models are not generalizable and hence not practical. A machine learning model trained with the population data may not yield accurate predictions for individuals who significantly differ in physiological characteristics. Therefore, we created a hybrid of generalized and personalized models to estimate PIs. With this aim, we used a clustering-based approach to group patients using their EDA features. Here, the subjects with similar EDA features were clustered, and NN-based PIs were constructed for each cluster separately. In this method, upon a new patient's arrival, we place the patient in the nearest cluster based on EDA signals and subsequently utilize the cluster-specific model to construct PIs. **Table 8** displays the number of individuals belonging to clusters, along with the PICP, MPIW, and NMPIW values of the PIs for each cluster. **Table 9** provides the details on PIs, including the average upper and lower bounds across pain levels, which are calculated for PICP values of 50%, 75%, 85%, and 95%.

Table 8: The hybrid model's results include PICP, MPIW, and NMPIW.

| Cluster | Number of Individuals | PICP | MPIW | NMPIW |
|---|---|---|---|---|
| 1 | 27 | 50% | 0.37 | 0.09 |
|   |    | 75% | 1.47 | 0.37 |
|   |    | 85% | 1.85 | 0.46 |
|   |    | 95% | 2.50 | 0.63 |
| 2 | 24 | 50% | 0.42 | 0.11 |
|   |    | 75% | 1.38 | 0.35 |
|   |    | 85% | 1.89 | 0.47 |
|   |    | 95% | 2.38 | 0.60 |
| 3 | 20 | 50% | 0.59 | 0.15 |
|   |    | 75% | 1.77 | 0.44 |
|   |    | 85% | 1.69 | 0.42 |
|   |    | 95% | 2.81 | 0.70 |
| 4 | 16 | 50% | 0.42 | 0.11 |
|   |    | 75% | 1.48 | 0.37 |
|   |    | 85% | 2.04 | 0.51 |
|   |    | 95% | 2.67 | 0.67 |

Table 9: The mean upper and lower bounds for each pain level averaged across clusters for PICP values of 50%, 75%, 85%, and 95%.

|  | HYBRID MODELS | | | | | | | |
|---|---|---|---|---|---|---|---|---|
|  | 50% PICP | | 75% PICP | | 85% PICP | | 95% PICP | |
| Target | Lower Bound | Upper Bound | Lower Bound | Upper Bound | Lower Bound | Upper Bound | Lower Bound | Upper Bound |
| 0 | 0.94 | 1.72 | 0.72 | 2.25 | 0.09 | 2.49 | 0.16 | 2.56 |
| 1 | 1.34 | 2.23 | 0.85 | 2.48 | 0.65 | 2.95 | 0.25 | 2.96 |
| 2 | 1.89 | 2.58 | 1.15 | 2.69 | 1.2 | 3.175 | 0.36 | 3.22 |
| 3 | 2.4 | 2.8 | 1.45 | 3.13 | 1.75 | 3.48 | 0.715 | 3.46 |
| 4 | 2.98 | 3.4 | 1.95 | 3.545 | 2.73 | 3.83 | 1.54 | 3.71 |

The number of subjects in clusters 1 through 4 are 27, 24, 20, and 16, respectively. Compared to generalized and personalized models, cluster-specific models perform better. The average MPIWs for clusters are 0.44, 1.52, 1.86, and 2.5 for 50%,75%,85%, and 95% PICP, respectively.

The pairwise Euclidian distance between the subjects in each cluster is calculated, and the distribution of distances is plotted in **Figure 4**. The average distances for clusters are 0.44, 0.46, 0.48, and 0.51, respectively. The average distances of clusters 1 and 2 are smaller than those of clusters 3 and 4. Smaller distances provide slightly better performance in constructing PIs for these clusters. However, overall, all these clusters perform similarly. **Figure 4** demonstrates that Cluster 4 has more outliers than the other clusters.

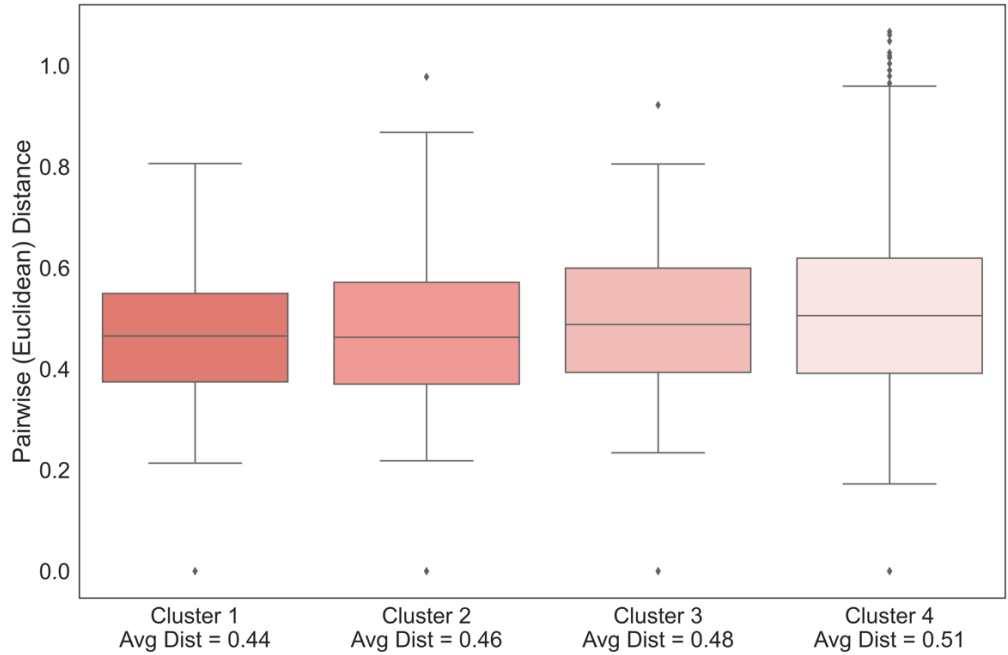

*Figure 4: The distribution of pairwise Euclidean distances for Clusters 1, 2, 3, and 4.*

To provide a clear picture, **Figure 5** illustrates the values of $MPIW_S$ and $PICP_S$ for each approach. The first four bars in the figure represent the interval width values of each cluster, followed by the fifth bar for the generalized model and the last for the personalized model. These bars are presented for the $PICP_S$ values of 50%, 75%, 85%, and 95%, respectively.

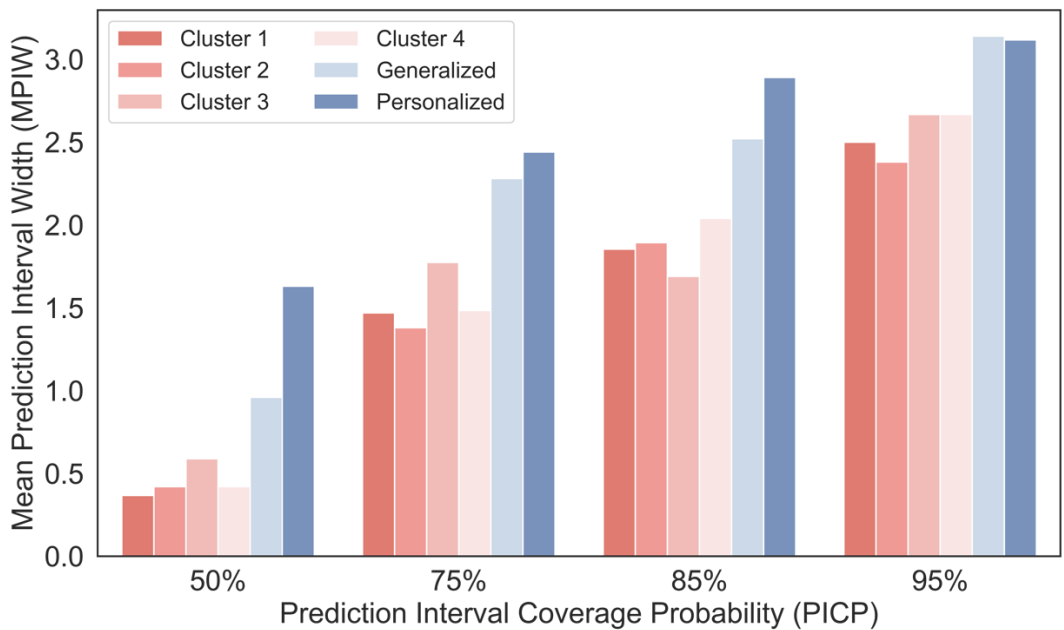

*Figure 5: The hybrid approach, which utilizes clustering techniques, outperforms other models and is considered a viable option for implementation in clinical settings.*

According to the results, using a clustering-based approach can significantly improve the estimation of PIs for pain intensity. Identifying subgroups of individuals who exhibit similar EDA patterns can enhance the quality and efficiency of constructing PIs. Such an approach is shown to be the most effective among various models and has practical applications in clinical settings.

## 5. Conclusion

In this work, we developed an NN-based prediction interval method to estimate pain intensity while capturing the prediction uncertainty. We used EDA signals from the BioVid Heat Pain Database of 87 individuals to develop and assess our models. We extracted 22 features from the EDA signals, including basic statistics of time-series values, stationarity, fits, entropy, physical nonlinear time-series analysis techniques, linear and nonlinear model parameters, linear correlations, and predictive power. We assessed the performance of our models using two primary metrics for prediction intervals: (1) accuracy, representing the confidence in our PI estimates as determined by PICP, and (2) dimension, reflecting resolution and quantified by PIW. We aim for a high PICP with a narrow PIW to ensure high-quality PIs.

We generated PIs with various NN-based PI estimation methodologies. First, we built a generalized model using $Loss_S$ and then compared the findings of the generalized model with those of the models trained using the $Loss_L$ and bootstrap approaches. The model using $Loss_S$ demonstrated superior performance compared to the $Loss_L$ and bootstrap-trained models, leading to reductions in PIW values of 22.4%, 7.9%, 16.7%, and 9.1% compared to the PIs generated by $Loss_L$, and 19.3%, 21.1%, 23.6%, and 26.9% compared to the PIs generated by bootstrap, across PICP values of 50%, 75%, 85%, and 95%, respectively. The findings indicate that $Loss_s$ outperformed the other two methods. Additionally, the results show a tradeoff between accuracy and dimension, whereby higher accuracy leads to a coarser dimension.

We then assessed $Loss_S$ performance on two other scenarios in addition to the generalized scenario. In the first case, we created a single model using all EDA signals from all subjects (population), which provides a generic model with reasonable performance and is valuable and applicable for pain intensity estimation in clinical settings. In the second case, we developed personalized models for individual subjects, though the training data was very limited for each subject. The personalized models are not generalizable to new subjects. Since we may have to create a new model for each new subject or patient, it is not practical in clinical settings. In the third case, we developed a cluster-based hybrid approach, where individuals were grouped based on the similarity of their EDA features, and a dedicated model was created for each cluster of subjects. This approach provides the highest quality PIs with improved accuracies and lower dimensions, with average prediction interval widths of 0.44, 1.52, 1.86, and 2.5 for the 50%, 75%, 85%, and 95% prediction interval coverage probabilities, respectively. Importantly, this approach is practical in clinical settings and allows the use of the same NN-based PIs model for a new pain-subject, eliminating the necessity to construct a unique model for every individual. In conclusion, the NN-based PI algorithm with $Loss_s$ effectively covers prediction uncertainty in pain intensity estimation. To our knowledge, this is the first study that estimated prediction intervals for pain intensity.

Data scarcity is a significant limitation in this study, particularly when developing personalized models. In future work, we will consider incorporating EMG, EEC, and video signals from individuals in the BioVid Heat Pain Dataset and training models using these additional data sources. Genetic algorithms for PI estimation are computationally complex, primarily due to their iterative nature. This complexity can become more pronounced when dealing with large datasets or high-dimensional optimization problems. Therefore, to overcome these challenges and enhance the efficiency of the optimization process, we will explore various optimization approaches, including particle swarm optimization, simulated annealing, and hybrid evolutionary algorithms.

In addition to its application in pain intensity estimation, the uncertainty quantification approach presented in this study holds promise for various other medical applications, including glucose level monitoring, blood pressure measurement, cardiovascular risk prediction, and drug dosage optimization. In each scenario, accurate and well-calibrated prediction intervals can significantly enhance the quality and reliability of medical decision-making, help manage patient expectations, and tailor interventions to individual patient needs.

# Supporting Information

**S1 Appendix:** Distribution, simple temporal statistics, linear and nonlinear autocorrelation, successive differences, and fluctuation analysis features are extracted from the time-series EDA signal [65].

| Time-Series Feature Category | Description |
|---|---|
| Distribution | Mode of z-scored distribution with a 5-bin histogram |
| | Mode of z-scored distribution with a 10-bin histogram |
| Simple temporal statistics | Longest period of consecutive values above the mean |
| | Time intervals between successive extreme events above the mean |
| | Time intervals between successive extreme events below the mean |
| Linear autocorrelation | First 1/e crossing of the autocorrelation function |
| | First minimum of autocorrelation function |
| | Total power in lowest fifth of frequencies in the Fourier power spectrum |
| | Centroid of the Fourier power spectrum |
| | Mean error from a rolling 3-sample mean forecasting |
| Nonlinear autocorrelation | Time-reversibility statistic, $\langle (x_{t+1} - x_t)^3 \rangle_t$ |
| | Auto mutual information, $m=2, \tau=5$ |
| | First minimum of the auto-mutual information function |
| Successive differences | Proportion of successive differences exceeding $0.04\sigma$ (Mietus et al. [74]) |
| | Longest period of successive incremental decreases |
| | Shannon entropy of two successive letters in equiprobable 3-letter symbolization |
| | Change in correlation length after iterative differencing |
| | Exponential fit to successive distances in 2-d embedding space |
| Fluctuation Analysis | Proportion of slower timescale fluctuations that scale with DFA (50% sampling) |
| | Proportion of slower timescale fluctuations that scale with linearly rescaled range fits |
| Others | Trace of covariance of transition matrix between symbols in the 3-letter alphabet |
| | Periodicity measure (Wang et al. [75]) |